%% file: acl_latex.tex
\pdfoutput=1

\documentclass[11pt]{article}

\usepackage[]{acl} 

\usepackage{times}
\usepackage{latexsym}

\usepackage[T1]{fontenc}

\usepackage[utf8]{inputenc}

\usepackage{microtype}

\usepackage{enumitem}

\usepackage{inconsolata}

\usepackage{graphicx}
\usepackage{amsmath}
\usepackage{multirow}
\usepackage{booktabs}
\usepackage{easyeqn}
\usepackage{bm}
\usepackage{amsfonts}
\usepackage{amssymb}
\usepackage{textcomp}
\usepackage[table]{xcolor}
\usepackage{colortbl} 
\definecolor{deepPurple}{RGB}{102, 0, 153}
\usepackage{listings}
\newcommand{\redtext}[1]{\textcolor{red}{#1}}

\newcommand{\vv}{\mathbf{v}}
\newcommand{\xv}{\mathbf{x}}
\newcommand{\yv}{\mathbf{y}}
\newcommand{\thetav}{\bm{\theta}}

\newcommand{\HC}{\mathcal{H}}

\usepackage[most]{tcolorbox}
\lstdefinestyle{plain}{
    basicstyle=\fontsize{9}{10}\ttfamily,
    keywordstyle=\color{blue},
    commentstyle=\color{gray},
    stringstyle=\color{green},
    showstringspaces=false,
    breaklines=true,
    breakatwhitespace=false,
    breakindent=0pt,
    escapeinside={(*@}{@*)}
}
\NewDocumentCommand{\yi}
{ mO{} }{\textcolor{magenta}{\textsuperscript{\textit{May}}\textsf{\textbf{\small[#1]}}}}
\NewDocumentCommand{\shujin}
{ mO{} }{\textcolor{cyan}{\textsuperscript{\textit{shujin}}\textsf{\textbf{\small[#1]}}}}
\NewDocumentCommand{\yuchen}
{ mO{} }{\textcolor{orange}{\textsuperscript{\textit{yuchen}}\textsf{\textbf{\small[#1]}}}}
\usepackage{hyperref} 
\usepackage{float}

\title{MMBoundary: Advancing MLLM Knowledge Boundary Awareness through Reasoning Step Confidence Calibration}

\author{Zhitao He$^{\textbf{1}}$ ~Sandeep Polisetty$^{\textbf{2}*}$ ~Zhiyuan Fan$^{\textbf{1}}$ ~Yuchen Huang$^{\textbf{1}}$ \\ \textbf{~Shujin Wu$^{\textbf{3}}$\thanks{Work done as a visiting student at HKUST.} ~Yi R. (May) Fung$^{\textbf{1}}$} \\
  $^{\textbf{1}}$Hong Kong University of Science and Technology\\
  $^{\textbf{2}}$UMass Amherst 
  ~$^{\textbf{3}}$University of Southern California \\
  \texttt{\{zhebu, yrfung\}@cse.ust.hk} \\}

\begin{document}
\maketitle
\input{sections/0_abstract1}
\begin{figure}[t]
  \includegraphics[width=\columnwidth]{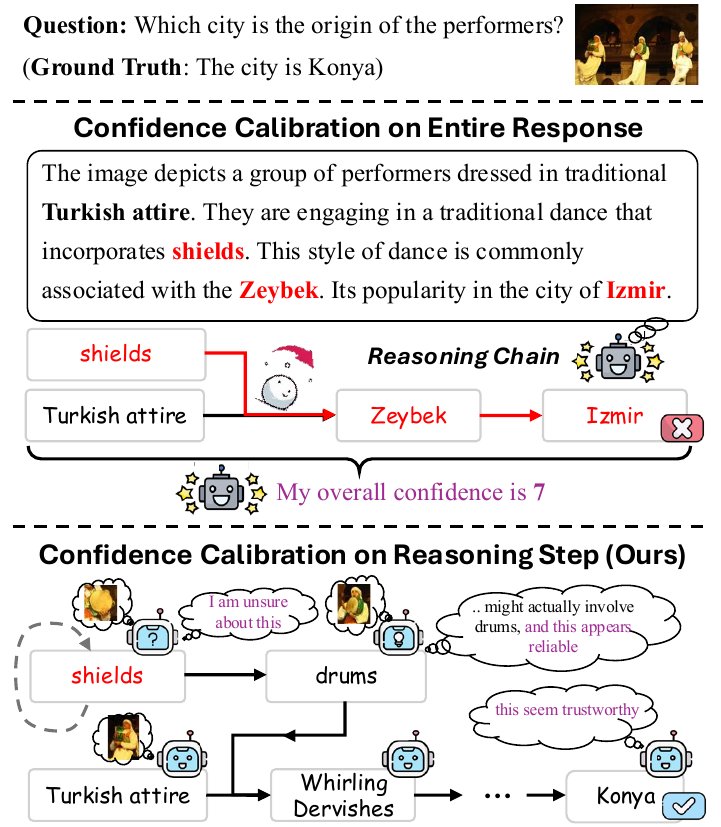}
  \caption{
  Confidence calibration on reasoning step enables MLLMs to express natural language confidence statements during inference, enhancing self-correction of low-confidence steps and ultimately reasoning toward correct answers. Traditional methods calibrate model confidence solely on entire response, which can lead to incorrect answers with high confidence. Due to space limitations, only the reasoning chain of our method is presented. The \redtext{red} and \textcolor{deepPurple}{purple} colors indicate incorrect knowledge and confidence estimates, respectively.} 
  \label{fig:snowballing}
\end{figure}
\input{sections/1_intro}

\begin{figure*}[t]
  \includegraphics[width=\textwidth]{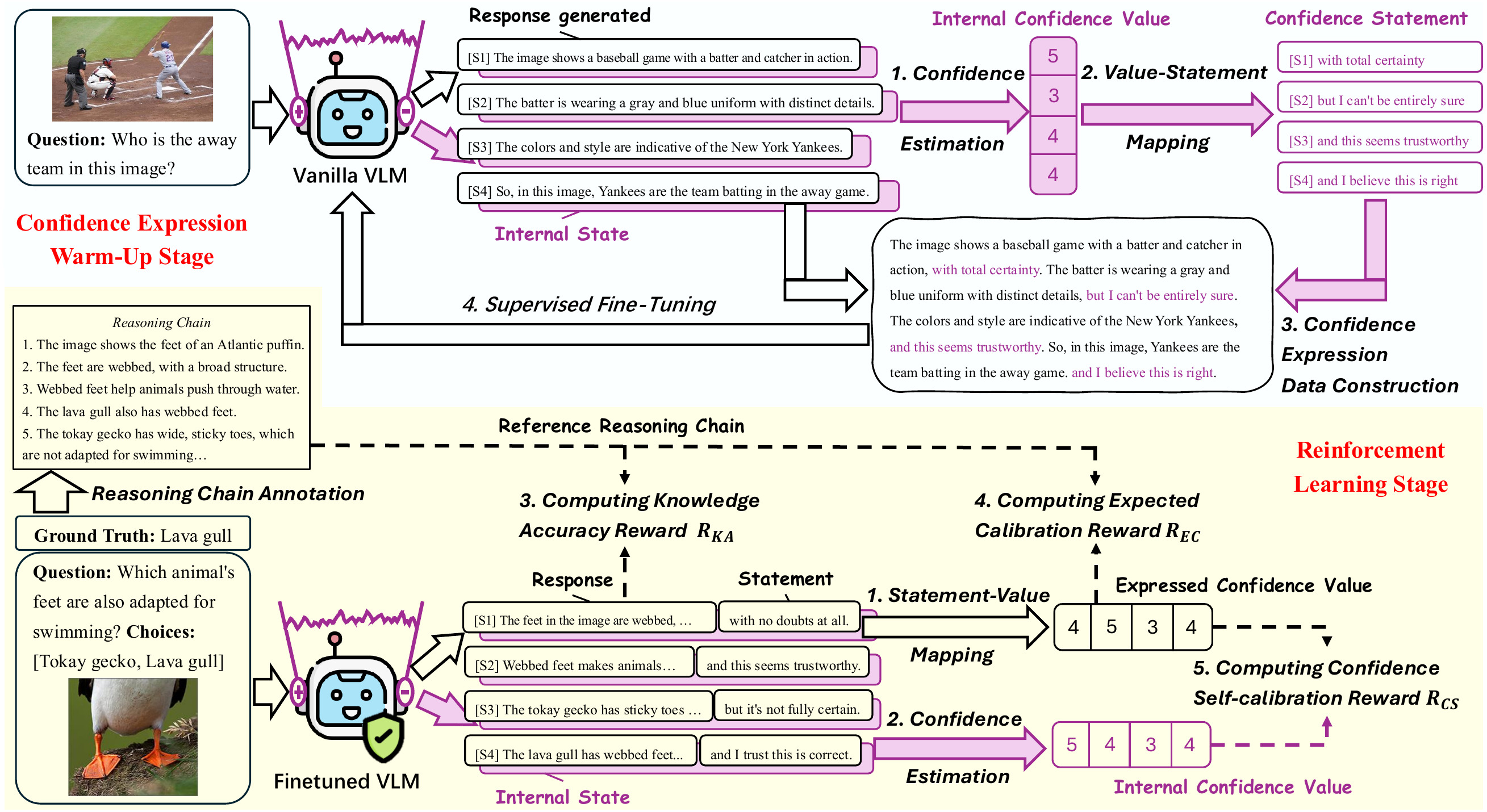}
  \caption{The overview of \textbf{MMBoundary}, which consists of two stages. The initial stage trains MLLMs via supervised learning to generate natural language confidence statement for each sentence, similar to human expression. The second stage employs reinforcement learning with three intuitively designed reward functions to further calibrate the expressed confidence estimates and enhance knowledge alignment.
  \textcolor[RGB]{150, 100, 200}{\rule{5pt}{5pt}} represents the internal states (i.e., the log probability of tokens) of model and the estimated internal confidence.} 
  \label{fig:MMB}
\end{figure*}

\input{sections/2_methodology}

\input{sections/4_experiment}

\input{sections/6_relatedwork}

\input{sections/7_conclusion}



\input{sections/8_limitations}

\bibliography{custom}

\input{sections/appendix}

\end{document}

%% file: sections/0_abstract1.tex
\begin{abstract}
In recent years, multimodal large language models (MLLMs) have made significant progress but continue to face inherent challenges in multimodal reasoning, which requires multi-level ({\em e.g.,} perception, reasoning) and multi-granular ({\em e.g.,} multi-step reasoning chain) advanced inferencing. 
Prior work on estimating model confidence tends to focus on the overall response for training and calibration, but fails to assess confidence in each reasoning step, leading to undesirable hallucination snowballing. 
In this work, we present \textbf{MMBoundary}, a novel framework that advances the knowledge boundary awareness of MLLMs through reasoning step confidence calibration. 
To achieve this, we propose to incorporate complementary textual and cross-modal self-rewarding signals to estimate confidence at each step of the MLLM reasoning process. 
In addition to supervised fine-tuning MLLM on this set of self-rewarding confidence estimation signal for initial confidence expression warm-up, we introduce a reinforcement learning stage with multiple reward functions for further aligning model knowledge and calibrating confidence at each reasoning step, enhancing reasoning chain self-correction. 
Empirical results show that MMBoundary significantly outperforms existing methods across diverse domain datasets and metrics, achieving an average of 7.5\% reduction in multimodal confidence calibration errors and up to 8.3\% improvement in task performance\footnote{Our code is publicly available at \url{https://github.com/Zhitao-He/MMBoundary}.}. 

\end{abstract}

%% file: sections/1_intro.tex
\section{Introduction}
Although multimodal large language models (MLLMs) demonstrate exceptional abilities in cross-modal reasoning, the reliability of their responses remains uncertain due to the inherent challenges of multimodal reasoning \cite{zhou2023analyzing, huang2024opera, chen2024unified,zhang2025vlm2benchcloserlookvlms}. In particular, erroneous knowledge can occur not only at the cross-modal reasoning level but also in the early stages of visual perception. However, MLLMs typically fail to explicitly indicate their uncertainty to avoid the propagation and amplification of knowledge errors \cite{liu2024survey, huang2024visual, bai2024hallucination, guan2024hallusionbench,huang2024pixels}. Therefore, it is crucial to enable MLLMs to accurately express confidence for each reasoning step during inference, enhancing reasoning chain self-correction.

Prior work on estimating model confidence tends to focus on the overall response for training and calibration
\cite{yang2023alignment, zhang2024r, lyu2024calibrating, xu2024sayself}. However, these methods fail to enable the trained models to express confidence estimates for different knowledge within generated content. As shown in Figure~\ref{fig:snowballing} (upper part), the trained MLLM generates incorrect information at the visual perception level ({\em i.e.,} misidentifying the "drum" as a "shield") without expressing its uncertainty, causing significant deviations in reasoning chain and ultimately producing an incorrect answer. Moreover, due to the logical coherence of the reasoning, the model still generates a high confidence score in its overall response. 


Therefore, in this work, we propose \textbf{MMBoundary}, a reinforced fine-tuning framework for advancing MLLM knowledge boundary awareness by reasoning step confidence calibration. 
Our method enables the model to express natural language confidence statement for each generated sentence, enhancing reasoning chain self-correction by scaling inference-time. Specifically, we introduce a confidence estimation module that integrates three effective text-based uncertainty methods—namely, length-normalized log probability, mean token entropy, and tokenSAR—and incorporates cross-modal constraint ({\em i.e.,} CLIPScore) to model the self-rewarding confidence signal from the perspective of its internal states. 
Then, we propose a mutual mapping between the detected score and predefined confidence statements to achieve two objectives: (1) by inserting confidence statements after the associated knowledge and training the model via supervised learning, we enable the model to naturally generate natural language statements for each sentence, similar to human expression; (2) by integrating internally detected confidence scores and those converted from model expressed statements into the reward modeling for reinforcement learning, we can achieve further confidence calibration, reducing the inaccuracy of model-expressed confidence. Moreover, we annotate the reference reasoning chain of training data to facilitate rigorous evaluation of MLLMs' knowledge at different reasoning levels, and incorporate model knowledge calibration into the reward modeling, encouraging MLLMs to faithfully express confidence while improving response quality.

Experimental results from both automatic and human evaluations across diverse domain datasets demonstrate that MMBoundary significantly reduces confidence calibration errors while simultaneously enhancing task performance. 

The contributions of our work can be summarized as follows: 
\begin{itemize}
    \item We present a novel framework, MMBoundary, for advancing the knowledge boundary awareness of multimodal language models through reasoning step confidence calibration.
    \item 
    We propose to integrate both textual and cross-modal self-rewarding signals for confidence estimation. Beyond supervised fine-tuning for initial confidence expression warm-up, we introduce a reinforcement learning stage with multiple reward functions to align model knowledge and calibrate confidence, enhancing self-correction in reasoning chain.
    \item Empirical results show that MMBoundary significantly outperforms existing methods, achieving an average reduction of 7.5\% in multimodal confidence calibration errors and up to 8.3\% improvement in task performance. 
\end{itemize}


%% file: sections/2_methodology.tex
\section{Problem Formulation}

Given a multimodal model $\bm{\pi}_{\bm{\theta}}$ with parameter $\bm{\theta}$, prior work focuses on enabling the model to output a confidence estimate for its entire response $\mathbf{y}$, formalized as:
\begin{equation}
\mathbf{y} = [z_1, z_2, \dots, z_T, c]
\end{equation}
Here, \(z_t\) represents the $t$-th generated sentence, \(c\) denotes the overall confidence estimate, \(T\) is the total number of sentences in the response. However, the trained model often assign high confidence incorrectly. Therefore we aim to train models to express fine-grained confidence estimate for each sentence during inference for enhancing reasoning chain self-correction. Thus, the output: 
\begin{equation}
\mathbf{y} = [z_1, c_1, z_2, c_2, \dots, z_T, c_T]
\end{equation} 
Each pair (\(z_t\), \(c_t\)) represents the $t$-th sentence generated by the model and its corresponding confidence statement, respectively. 

\section{Methodology}

Our framework consists of two stages: the confidence expression warm-up stage and the reinforcement learning stage, as shown in Figure \ref{fig:MMB}. 

\subsection{Confidence Expression Warm-Up}
\label{stage1}
\subsubsection{Internal Confidence Estimation}
\label{sec:ice}
In this section, we propose to leverage multiple text-based uncertainty methods and incorporate visual constraint to estimate MLLM's confidence. Previous work primarily relies on model response consistency as a confidence indicator. However, these methods fail to assess confidence across distinct knowledge in generated content and do not consider the correlation between the response and the visual information, limiting their applicability in multimodal scenarios. Drawing on recent research \cite{xiao2022uncertainty, fadeeva2023lm, vashurin2024benchmarking}, we utilize the following efficient and effective uncertainty estimation methods to create our confidence indicator:

(1) \textbf{\textit{Length-normalized log probability}} calculates the average negative log probability of the tokens generated:
\begin{equation}
U_\mathrm{LNLP}(\yv, \xv; \thetav) = 
\exp\Biggl\{
    -\frac{1}{L} 
    \log P(\yv \mid \xv, \thetav)
\Biggr\},
\label{eq:nsp}
\end{equation}
where \( \xv \) denotes the input, \( \yv \) denotes the output, and \( \thetav \) represents the model parameters.

(2) \textbf{\textit{Mean token entropy}} \cite{fomicheva2020unsupervised} computes the average entropy for each token in the generated sentence:
  \begin{equation}
    U_{MTE}(\yv, \xv; \thetav) = \frac{1}{L} \sum\nolimits_{l = 1}^L \HC(y_l \mid \yv_{<l}, \xv, \thetav),
  \label{eq:entropy}
  \end{equation}
where $\HC(y_l \mid \yv_{<l}, \xv, \thetav)$ is an entropy of the token distribution $P(y_l \mid \yv_{<l}, \xv, \thetav)$.
  
(3) \textbf{\textit{TokenSAR}}~\cite{duan2024shifting} computes the weighted average of the negative log probability of generated tokens, considering their relevance to the entire generated text. For a given sentence similarity function $g(\cdot, \cdot)$ and token relevance function $R_T(y_k, \yv, \xv) = 1 - g(\xv \cup \yv, \xv \cup \yv \setminus y_k)$, the resulting estimate is computed as:
\begin{multline}
U_\mathrm{TokenSAR}(\yv, \xv; \thetav) = \\
-\sum\nolimits_{l = 1}^L \tilde{\mathrm{R}}_T(y_l, \yv, \xv) \log P(y_l \mid \yv_{<l}, \xv, \thetav), 
\label{eq:tsar}
\end{multline}
where $\tilde{\mathrm{R}}_T(y_k, \yv, \xv) = \frac{\mathrm{R}_T(y_k, \yv, \xv)}{\sum\nolimits_{l = 1}^L \mathrm{R}_T(y_l, \yv, \xv)}$.\\

(4) \textbf{\textit{CLIPScore}}~\cite{hessel2021clipscore} evaluates the relevance between the generated sentence and input image. Since CLIP's vision encoder aligns with the target MLLM's, we employ CLIPScore to represent the sentence-image uncertainty. For an image with visual CLIP embedding $v$ and a sentence with textual CLIP embedding $s$:
\begin{figure}[htbp]
  \includegraphics[width=\columnwidth]{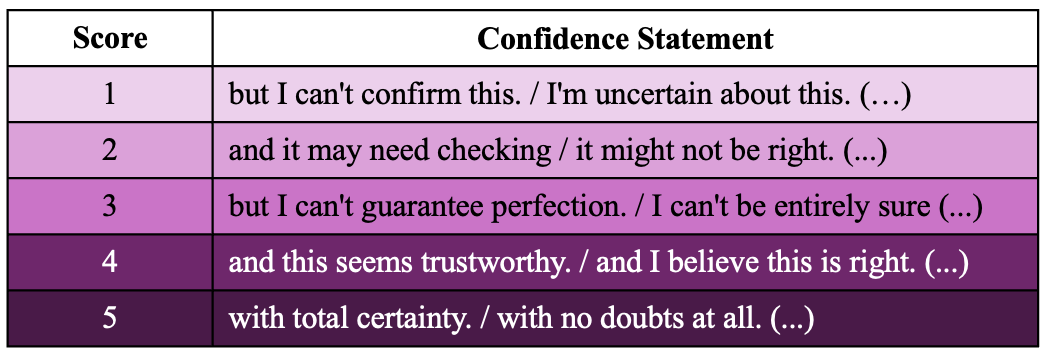}
  \caption{We preset a confidence statement pool for each confidence score. The five levels correspond to uncertain, slightly uncertain, moderately confident, highly confident, and fully confident. More statements are shown in Appendix ~\ref{sec:mapping_table}. } 
  \label{fig:mapping}
\end{figure}
\begin{equation}
    U_\mathrm{CLIPScore}(\vv, \mathbf{s}) = \max\left(\cos(\vv, \mathbf{s}), 0\right)
\label{eq:clipscore}
\end{equation}
We normalize \( U_\mathrm{i} \) across the entire dataset using min-max normalization to ensure their values are within the range [0, 1]. Then, we compute the final weighted average as:
\begin{multline}
U_\mathrm{Final} = w_0 U_{\mathrm{LNLP}} + w_1 U_{\mathrm{MTE}} \\
+ w_2 U_{\mathrm{TokenSAR}} + w_3 U_{\mathrm{CLIPScore}}
\end{multline}
where \(w_i\) are the respective weights for each component. The closer \( U_\mathrm{Final} \) is to 0, the greater the certainty of the model. Then, we use the distribution of \( U_\mathrm{Final} \) to define confidence levels for the model, considering the uneven distribution of \( U_\mathrm{Final} \). Confidence levels \( C_\mathrm{v} \) from 5 to 1 correspond to the intervals of \( U_\mathrm{Final} \) as [0, $\mu$ - $\sigma$, $\mu$ + $\sigma$, $\mu$ + 2$\sigma$, $\mu$ + 3$\sigma$, 1], with higher confidence levels indicating greater model confidence. Here, $\mu$ and $\sigma$ represents the average and the standard deviation of \( U_\mathrm{Final} \). 
We further validate the effectiveness of this confidence level classification method in Section \ref{anaylsis_CE}.

\subsubsection{Confidence Score-Statement Mapping}

This module, as shown on the right side of Figure~\ref{fig:MMB}, aims to establish a mutual mapping between the detected score and predefined confidence statements. 
First, we construct statement pools for each confidence level, as shown in Figure \ref{fig:mapping}. These statements can be naturally appended to the end of sentences, providing a concise expression of the model's confidence estimates, similar to human expression. During the \textit{Confidence Expression Warm-Up Stage}, we randomly select statements from the corresponding pools based on the detected scores and insert them into the model's original response to create data for fine-tuning. In the \textit{Reinforcement Learning Stage}, after obtaining the confidence statements from the model's output, we encode these statements into vectors using an encoder model\footnote{\url{https://huggingface.co/sentence-transformers/all-MiniLM-L6-v2}} and compute the cosine similarity with all embeddings in the different confidence pools to achieve reverse mapping of statements to confidence scores.

\subsubsection{Supervised Fine-Tuning}
Specifically, the model undergoes fine-tuning on our constructed data $\mathcal{D}$ consisting of tuples: \((\mathbf{x}, \mathbf{y})\), where the input \(\mathbf{x}\) comprises an image \(I\) and a question \(Q\). 
At step \(s_t\), the sentence with its confidence statement (\(z_t\), \(c_t\)) are generated by the model's policy $\bm{\pi}_{\bm{\theta}}$. The next state \(s_{t+1}\) is:
\begin{equation}
s_{t+1} =
\begin{cases} 
x, & t = 0 \\
[s_t, z_t, c_t], & 1 \leq t \leq T
\end{cases}
\label{eq:function_label}
\end{equation}
We fine-tune the vanilla model via supervised learning. The loss function can be written as:
\begin{equation} 
\mathcal{L}_{FT}(\theta) = -\mathbb{E}_{(\mathbf{x}, \mathbf{y}) \sim \mathcal{D}} \left[ \sum_{t=1}^T \log \pi_\theta(z_t, c_t | s_t) \right]
\end{equation}

\subsection{Reinforcement Learning} 
\label{stage2}
As noted by \citealp{xu2024sayself}, the model undergoing supervised training tends to generate uniform confidence levels, which may impact task performance. Therefore, we employ reinforcement learning with reward signals involving model knowledge alignment, internal confidence and external confidence calibration to encourage model to faithfully express confidence while simultaneously improving the quality of responses. Specifically, we sample questions from the training data and prompt the model to generate responses. 

\noindent \textbf{(1) Knowledge Accuracy Reward} evaluates whether the knowledge in generated response is aligned with annotated reference chain, thereby ensuring the reliability of the generated content. 
Specifically, if the $t$-th generated sentence $z_t$ matches the knowledge in reference chain, $R_{KA_t}$ is 1. Refer to the "Step Matched" example in Figure~\ref{fig:MECE}. After evaluating all generated sentences, the reward is normalized:
\begin{equation} 
R_{KA} = \frac{1}{T} \sum_{t=1}^{T} R_{KA_t}
\end{equation}
where $T$ is the total number of sentences.

\noindent \textbf{(2) Expected Calibration Reward} is consistent with \citet{xu2024sayself}, but we extend it to sentence-level. This reward function measure the correlation between the expressed confidence and the ground truth. The Expected Calibration Reward (\(R_{EC}\)) is fundamentally consistent with Expected Calibration Error (ECE). We expect the model's confidence scores to properly reflect answer quality, i.e. lower confidence for poor-quality answers and vice versa.
The reward function is formalized as follows:
\begin{equation} 
R_{EC} = \frac{1}{T} \sum_{t=1}^{T} [1 - 2 \cdot \left(\mathbb{I}(z_t) - \mathrm{EV}(c_t)\right)^2]
\end{equation}
where \( \mathbb{I}(\cdot) \) is the indicator function that returns 1 if the sentence is correct compared with reference chain, and 0 otherwise. \(\mathrm{EV(c_t)}\) represents the expressed confidence score, which is obtained by mapping and normalizing the confidence statements generated by the model. The confidence score is normalized between 0 and 1. 

\noindent \textbf{(3) Confidence Self-Calibration Reward} is based on the consistency between the expressed confidence and internal confidence of MLLMs:
\begin{equation} 
R_{\text{CS}} = \frac{1}{T} \sum_{t=1}^{T} [1-2 \cdot \left(\mathrm{IV}(z_t) - \mathrm{EV}(c_t)\right)^2]
\end{equation}
where $\mathrm{IV(z_t)}$ represents the internal confidence score, which is estimated by our method in Secion~\ref{sec:ice}. This reward encourages the model to express its confidence level as accurately as possible, aligning its external expression with internal belief. 
Thus, the overall reward for response is:
\begin{equation} 
R =  \alpha R_{KA} + \beta R_{EC} + \gamma R_{CS}
\end{equation}

Lastly, we employ the Proximal Policy Optimization (PPO) algorithm \cite{schulman2017proximal} for training. The model's policy objectives is:
\begin{equation}
\begin{split}
\mathcal{L}_{RL}(&\boldsymbol{\theta})  = -\mathbb{E_{\mathbf{y} \sim \boldsymbol{\pi}_{\theta_{\text{old}}}}}\Bigg[ \min \Bigg( \frac{\boldsymbol{\pi}_{\boldsymbol{\theta}} (z_t, c_t | s_t)}{\boldsymbol{\pi}_{\theta_{\text{old}}} (z_t, c_t | s_t)} \hat{A}_t, \\
&\text{clip}\left( \frac{\boldsymbol{\pi}_{\boldsymbol{\theta}} (z_t, c_t | s_t)}{\boldsymbol{\pi}_{\theta_{\text{old}}} (z_t, c_t | s_t)}, 1-\epsilon, 1+\epsilon \right) \hat{A}_t \Bigg) \Bigg]
\end{split}
\label{eq:loss_policy}
\end{equation}
The advantage estimate \( \hat{A}_t \) \cite{schulman2015high} is derived by calculating the difference between the anticipated future rewards under the current policy and the baseline or value function. Implementation and data details can be found in Appendix \ref{sec:training}.

\begin{table*}[h!]
\centering
\resizebox{\linewidth}{!}{
\begin{tabular}{l|cccc|cccc|cccc}
\toprule
\multirow{2}{*}{\textbf{Model}} & \multicolumn{4}{c}{\textbf{A-OKVQA}} & \multicolumn{4}{c}{\textbf{ScienceVQA}} & \multicolumn{4}{c}{\textbf{CulturalVQA}}  \\ \cmidrule(lr){2-13} & \textbf{ECE} ($\downarrow$) & \textbf{MECE} ($\downarrow$) & \textbf{Acc} ($\uparrow$)  & \textbf{F1} ($\uparrow$) & \textbf{ECE} ($\downarrow$) & \textbf{MECE} ($\downarrow$) & \textbf{Acc} ($\uparrow$) & \textbf{F1} ($\uparrow$) & \textbf{ECE} ($\downarrow$) & \textbf{MECE} ($\downarrow$) & \textbf{Acc} ($\uparrow$) & \textbf{F1} ($\uparrow$)   \\ \midrule
DPV & 0.563 & 0.582 & 0.650 & 0.512 & 0.593 & 0.611 & 0.582 & 0.414 & 0.624 & 0.650 & 0.334 & 0.482 \\
DPS & 0.511 & 0.554 & 0.675 & 0.535 & 0.574 & 0.575 & 0.575 & 0.427 & 0.572 & 0.594 & 0.354 & 0.485\\
SC & 0.435 & 0.492 & 0.701 & 0.548 & 0.463 & 0.534 & 0.602 & 0.442 & 0.491 & 0.554 & 0.371 & 0.511 \\
Multisample & 0.413 & 0.430 & 0.683 & 0.543 & 0.446 & 0.500 & 0.596 & 0.434 & 0.463 & 0.505 & 0.362 & 0.538 \\
SaySelf & 0.345 & 0.384 & 0.734 & 0.603 & 0.386 & 0.462 & 0.633 & 0.483 & 0.375 & 0.437 & 0.417 & 0.571 \\
Conf-CSR & 0.408 & 0.437 & 0.785 & 0.618 & 0.453 & 0.503 & 0.694 & 0.502 & 0.472 & 0.513 & 0.435 & 0.582 \\
RCE & 0.361 & 0.394 & 0.788 & 0.620 & 0.413 & 0.475 & 0.671 & 0.497 & 0.408 & 0.453 & 0.412 & 0.577 \\
DRL & 0.395 & 0.453 & 0.746 & 0.614 & 0.476 & 0.513 & 0.654 & 0.485 & 0.453 & 0.502 & 0.392 & 0.564 \\

\rowcolor{cyan!3.5}
MMBoundary & \underline{\textbf{0.316}} & \underline{\textbf{0.304}} & 0.835 & \underline{\textbf{0.661}} & 0.354 & \underline{\textbf{0.392}} & 0.703 & \underline{\textbf{0.565}} & \underline{\textbf{0.337}} & \underline{\textbf{0.361}} & \underline{\textbf{0.448}} & \underline{\textbf{0.665}} \\
\hline
\hline
\multicolumn{13}{c}{\textbf{Warm-Up Stage}} \\
\hline
\rowcolor{cyan!3.5}
w/o $\text{U}_{LNLP}$  & 0.327 & 0.343 & 0.815 & 0.642 & 0.369 & 0.421 & 0.698 & 0.548 & 0.356 & 0.397 & 0.423 & 0.639 \\
\rowcolor{cyan!3.5}
w/o $\text{U}_{MTE}$  & 0.337 & 0.332 & 0.824 & 0.653 & 0.385 & 0.441 & 0.682 & 0.536 & 0.349 & 0.378 & 0.430 & 0.643 \\
\rowcolor{cyan!3.5}
w/o $\text{U}_{TSAR}$  & 0.324 & 0.358 & 0.813 & 0.627 & 0.352 & 0.417 & 0.694 & 0.546 & 0.361 & 0.403 & 0.438 & 0.655\\
\rowcolor{cyan!3.5}
w/o $\text{U}_{CLIPS}$  & 0.337 & 0.354 & 0.806 & 0.631 & 0.372 & 0.435 & 0.681 & 0.532 & 0.358 & 0.394 & 0.426 & 0.637 \\
\rowcolor{cyan!3.5}
w $\text{U}_{\text{Max}}$  & 0.362 & 0.386 & 0.774 & 0.583 & 0.391 & 0.467 & 0.663 & 0.494 & 0.378 & 0.425 & 0.403 & 0.589\\
\rowcolor{cyan!3.5}
w/o $\text{S-S}_{Mapping}$  & 0.340 & 0.362 & 0.793 & 0.634 & 0.377 & 0.443 & 0.684 & 0.531 & 0.365 & 0.398 & 0.427 & 0.602 \\
\hline
\multicolumn{13}{c}{\textbf{Reinforcement Learning Stage}} \\
\hline
\rowcolor{cyan!3.5}
w/o $\text{R}_{KA}$  & 0.325 & 0.347 & 0.802 & 0.629 & \underline{\textbf{0.334}} & 0.410 & 0.686 & 0.535 & 0.348 & 0.370 & 0.437 & 0.632\\
\rowcolor{cyan!3.5}
w/o $\text{R}_{EC}$  & 0.332 & 0.357 & 0.819 & 0.635 & 0.363 & 0.426 & \underline{\textbf{0.712}} & 0.548 & 0.359 & 0.392 & 0.422 & 0.648 \\
\rowcolor{cyan!3.5}
w/o $\text{R}_{CS}$  & 0.343 & 0.368 & \underline{\textbf{0.857}} & 0.648 & 0.372 & 0.449 & 0.693 & 0.556 & 0.368 & 0.417 & 0.436 & 0.640\\

\rowcolor{cyan!3.5}
w/o $\text{RL}$  & 0.392 & 0.427 & 0.768 & 0.581 & 0.419 & 0.481 & 0.663 & 0.495 & 0.408 & 0.456 & 0.419 & 0.595\\
\bottomrule
\end{tabular}}
\caption{The evaluation results of models and various ablations of our framework. CulturalVQA is the out-of-distribution dataset. 
\textbf{w/o $\text{U}_{LNLP}$}, \textbf{w/o $\text{U}_{MTE}$}, \textbf{w/o $\text{U}_{TSAR}$}, and \textbf{w/o $\text{U}_{CLIPS}$} represent MMBoundary without the three text-based uncertainty estimation methods and visual information uncertainty estimation, respectively; \textbf{w $\text{U}_{Max}$} indicates the confidence determined using the max pooling method from the four uncertainty estimation scores; \textbf{w/o $\text{S-S}_{Mapping}$} denotes MMBoundary without confidence score-statement mapping; \textbf{w/o $\text{R}_{KA}$}, \textbf{w/o $\text{R}_{EC}$}, and \textbf{w/o $\text{R}_{CS}$} represent MMBoundary without knowledge accuracy reward, expected calibration reward, and confidence self-calibration reward, respectively; \textbf{w/o $\text{RL}$} denotes MMBoundary without reinforcement learning. 
}
\label{tab:results_AE}
\end{table*}

\begin{table*}[h!]
\centering
\resizebox{\linewidth}{!}{
\begin{tabular}{l|cccc|cccc|cccc}
\toprule
\multirow{2}{*}{\textbf{Model}} & \multicolumn{4}{c}{\textbf{A-OKVQA}} & \multicolumn{4}{c}{\textbf{ScienceVQA}} & \multicolumn{4}{c}{\textbf{CulturalVQA}}  \\ 
\cmidrule(lr){2-13}
 & \textbf{Faithful} & \textbf{Concise} & \textbf{Granular} & \textbf{Avg.} & \textbf{Faithful} & \textbf{Concise} & \textbf{Granular} & \textbf{Avg.} & \textbf{Faithful} & \textbf{Concise} & \textbf{Granular} & \textbf{Avg.} \\ \midrule
Multisample & 4.20 & 5.17 & 4.06 & 4.47 & 4.77 & 5.24 & 4.53 & 4.85 & 3.91 & 5.63 & 4.72 & 4.75\\
SaySelf & 7.28 & \underline{\textbf{7.49}} & 6.47 & 7.08 & 7.49 & 7.18 & 6.28 & 6.98 & 7.12 & 6.81 & 6.58 & 6.83\\ 
Conf-CSR & 6.47 & 5.73 & 5.82 & 6.01 & 6.74 & 5.61 & 6.40 & 6.23 & 6.38 & 5.45 & 5.86 & 5.89\\ 
RCE & 6.73 & 6.58 & 7.41 & 6.90 & 7.55 & 6.92 & 7.12 & 7.19 & 6.84 & 6.19 & 6.82 & 6.62\\
DRL & 6.54 & 6.13 & 6.95 & 6.54 & 6.81 & 5.97 & 6.34 & 6.37 & 6.55 & 5.63 & 6.07 & 6.08\\
\rowcolor{cyan!3.5}
MMBoundary & \underline{\textbf{7.83}}   &7.25   &\underline{\textbf{8.18}} &\underline{\textbf{7.75}}   &\underline{\textbf{8.35}}   &\underline{\textbf{7.46}}&\underline{\textbf{8.02}}   &\underline{\textbf{7.94}}   &\underline{\textbf{7.66}} & \underline{\textbf{7.17}} & \underline{\textbf{8.26}} & \underline{\textbf{7.69}}\\ 
\bottomrule
\end{tabular}}
\caption{The human evaluation results of strong baselines and our framework. We provide a panel of three graduate students with 50 random entries from each setting, asking them to
evaluate whether each entry meets the criteria (\textbf{Faithful}, \textbf{Concise}, \textbf{Granular}) and to give a score from 1 to 10. The final result is the average score.} 
\label{tab:results_Human}
\end{table*}


%% file: sections/4_experiment.tex
\section{Experiments}

\subsection{Dataset}
In order to evaluate the model’s robustness and generalizability across diverse scenarios, we select the following datasets from different domains: \textbf{A-OKVQA} \cite{schwenk2022okvqa}, a general domain dataset designed to evaluate models on complex visual question answering tasks involving multi-hop reasoning, commonsense understanding, and external knowledge integration; \textbf{ScienceVQA} \cite{lu2022learn}, a large-scale multimodal dataset designed for science question answering, featuring questions across natural science, social science, and language science; \textbf{CulturalVQA} \cite{nayak2024benchmarking}, a visual question-answering benchmark evaluating MLLMs on understanding geo-diverse cultural concepts beyond general scene understanding.

\subsection{Reasoning Chain Annotation}
To simultaneously calibrate the model's knowledge and confidence levels, we conduct detailed reasoning chain annotations for each question in the training dataset. As shown in Figure \ref{fig:data_construct}, for each question, we prompt the GPT-4o to generate analysis (inference chain) structured in the perception and reasoning level. The former identifies key visual elements in the image that are most relevant to the question and answer, while the latter provides granularity reasoning that justifies why the answer is correct. Each level should include concise, interconnected sentences, with each sentence conveying a single piece of knowledge. Then, we perform filtering and quality evaluation to ensure the accuracy and consistency. Due to space limitations, please refer to Appendix \ref{sec:annotation} for more details.

\subsection{Evaluation Metrics}

Consistent with previous research \cite{chen2022close, xu2024sayself}, we evaluate models from three perspectives using six different metrics:

\noindent \textbf{(1) Confidence Calibration Performance}: We adopt 3 calibration metrics. First, we use the Expected Calibration Error (\textbf{ECE}) score \cite{guo2017calibration}. Then, we extend the ECE score to measure the confidence calibration error of each knowledge within reasoning chain, which we refer to as \textit{Multi-granularity Expected Calibration Error} (\textbf{MECE}). The MECE score evaluates the correlation between the confidence estimates expressed in generated sentences and their corresponding correctness, as shown in Figure~\ref{fig:MECE}. Details of MECE computation process is in the Appendix \ref{metric:MECE}. For all responses $A$ generated by MLLMs:
\begin{equation}
\text{MECE} = \frac{1}{|A|} \sum_{a \in A} \frac{1}{|a|} \sum_{(z,c) \in a} \left| \mathbb{I}(z) - \text{conf}(c) \right|
\end{equation}
Here, $a$ represents the model's response to the question, while $z$ and $c$ represent the sentences in the response and their corresponding confidence statements, respectively. $\text{Conf}(\cdot)$ represents the numerical value of the confidence statement.

\noindent \textbf{(2) Task Performance}: We adopt 2 metrics. First, we measure the typical \textit{\textbf{Accuracy}}. Second, to identify model responses containing erroneous knowledge and mitigate the risk of them being assigned high confidence, we evaluate the quality of the model's reasoning chain by employing the metric in 
\textit{\textbf{Reasoning Chain F1}} score \cite{ho2022wikiwhy}. This metric compares the information contained in the predictions and references. 
We present implementation details in the Appendix \ref{metric:F1}.

\noindent \textbf{(3) Human Evaluation}: 
Automated model evaluation may not accurately capture the subtle differences between different responses \cite{goyal2022news,ho2022wikiwhy,he2023lego}. Therefore, we conduct additional manual evaluation. We provide a panel of three graduate students with 50 random entries from each setting, asking them to evaluate whether each entry meets the following criteria and to give a score from 1 to 10, consistent with \citep{xu2024sayself}. 1) \textit{\textbf{Faithful}}: whether the response faithfully expresses the confidence; 2) \textit{\textbf{Concise}}: whether the response conveys necessary information clearly and without excess; 3) \textit{\textbf{Granular}}: whether the response contains confidence estimates for distinct knowledge. The final result is the average score of these criteria. 

\subsection{Baselines}

We compare with the following methods: (1) \textbf{DPV}: directly prompting the vanilla MLLMs to give a response with a confidence score; (2) \textbf{DPS}: direct prompting the vanilla MLLMs to give a response with a confidence statement; (3) \textbf{SC} \cite{xiong2023can}: deriving the confidence estimates of MLLMs based on diverse sampling; 
(4) \textbf{Multisample} \cite{yang2023alignment}: training MLLMs to generate confidence estimates that align with the confidence derived from self-consistency; (5) \textbf{SaySelf} \cite{xu2024sayself}: analyzing inconsistencies in multiple sampled responses, with the resulting data used for supervised fine-tuning and then confidence estimates calibrated through reinforcement learning based on task supervision; (6) \textbf{Conf-CSR}: converting the calibrated self-reward \cite{zhou2024calibrated} of each sentence into the model's confidence score and utilize DPO \cite{rafailov2024direct} for optimization; (7) \textbf{RCE}: training the model to first generate a complete response and then produce confidence estimates for each sentence; (8) \textbf{DRL}: directly employing our reinforcement learning method to train model. We use LLaVA-NEXT 7B \cite{liu2024llava} as backbone model for all methods to ensure a fair comparison. 
To prove that our method can generalize across models, we also conduct experiments on Qwen2VL 7B \cite{wang2024qwen2} in Appendix \ref{sec:extra_experiment}. 

\subsection{Main Experimental Results}

\noindent \textbf{Confidence Calibration Performance.} We present the ECE and MECE results in Table \ref{tab:results_AE} and the AUROC results in Appendix (Table \ref{tab:results_auroc}), which measure the correlation between the expressed confidence and the ground truth. 
The findings indicate that MMBoundary outperforms other methods in reducing confidence calibration errors and enhancing the ability to distinguish confidence between correct and incorrect answers (AUROC). This conclusion is validated on both in-distribution datasets (AOKVQA and ScienceVQA, with a 7.5\% improvement in the MECE score and out-of-distribution datasets (CulturalVQA, showing an increase of 6.6\%), highlighting the generality of our framework. Compared to baseline methods (DRL, RCE, and SaySelf) that also require annotated data during the RL phase, our method achieves the best performance, improving MECE and F1 scores by 5.2\% and 7.8\%, respectively. When no annotated data is used, our method (w/o RL) surpasses the baselines (Multisample and Conf-CSR) by up to 5.7\% (MECE) under the same settings.

\noindent \textbf{Task Performance.} We comprehensively evaluate the task performance of the model using the final answer accuracy and the Reasoning Chain F1 score, as presented in Table \ref{tab:results_AE}. The results show that our method surpasses other baselines across three datasets, achieving up to 8.3\% improvement in CulturalVQA. Unlike Conf-CSR and SaySelf, which rely solely on task-oriented reward or the expected calibration reward, our approach integrates knowledge alignment along with internal and external confidence calibration into reward modeling. The results demonstrate that our framework improves the model's knowledge boundary awareness while simultaneously enhancing its task performance. We conduct paired t-tests on the experimental results of MMBoundary, showing significant advantages over the baselines (p-value < 0.05). 

\noindent \textbf{Human Evaluation.} We conduct human evaluation of the responses generated by our method and other strong baselines across the dimensions of faithfulness, conciseness, and granularity, with the results shown in Table \ref{tab:results_Human} and Figure \ref{fig:eval_box}. We observe that our framework demonstrates statistically significant improvements over three dimensions. SaySelf performs well in the concise dimension for content, but it is designed only to estimate confidence for the entire response, lacking the ability to generate confidence for each step of the reasoning process. We perform a Kappa test on the faithfulness evaluation results to assess inter-annotator agreement, obtaining a Kappa value of 0.79. 

\begin{table}[t]
\small
\centering
\begin{tabular}{c|cccc}
\toprule
Dateset & Mean & Var & Std & <0.1   \\ \midrule
A-OKVQA & 0.0443  & 0.0015 &  0.0397 & 96\% \\
ScienceVQA & 0.0578 & 0.0014 & 0.0374 & 93\% \\
CulturalVQA & 0.0522 & 0.0015 & 0.0397 & 94\%  \\ 
\bottomrule
\end{tabular}
\caption{Comparison between our internal confidence estimation (ICE) and widely adapted self-consistency-based estimation (SCE). We compute \( |C_{\text{ICE}} - C_{\text{SCE}}| \) to demonstrate the correlation between the two methods.} 
\label{tab:ICE}
\end{table}

\section{Discussion}
\subsection{Influence of Different Components}
We conduct an extensive ablation study to verify the effectiveness of different components, with results shown in Table \ref{tab:results_AE}. Compared to the version without RL, supervised fine-tuning enable model to express confidence during inference. Incorporating RL with our reward signals further improves confidence precision, with an average 9.2\% increase in the MECE score. Each reward term encourages the model to focus on distinct aspects (e.g., knowledge alignment or confidence calibration). When $R_{KA}$ is removed (leaving only confidence calibration rewards), the model prioritizes confidence calibration, leading to improved Expected Calibration Error performance on ScienceVQA and resulting in an average performance drop of 3.1\%. Conversely, removing $R_{EC}$ or $R_{CS}$ shifts the model's focus toward knowledge alignment (as the weight of knowledge alignment rewards becomes dominant), thereby improving accuracy (Acc) on A-OKVQA. The removal of $R_{EC}$ and $R_{CS}$ leads to a maximum decrease of 6.4\% in the confidence calibration performance. Furthermore, the results show that all four selected uncertainty estimation methods enhance model performance, with the $U_{TSAR}$ having the most significant impact on confidence calibration. Moreover, the conversion between confidence scores and statements ($\text{S-S}_{Mapping}$) positively impacts the model's confidence calibration, resulting in an average improvement of 4.8\%. 

\begin{figure}[t]
  \includegraphics[width=\columnwidth]{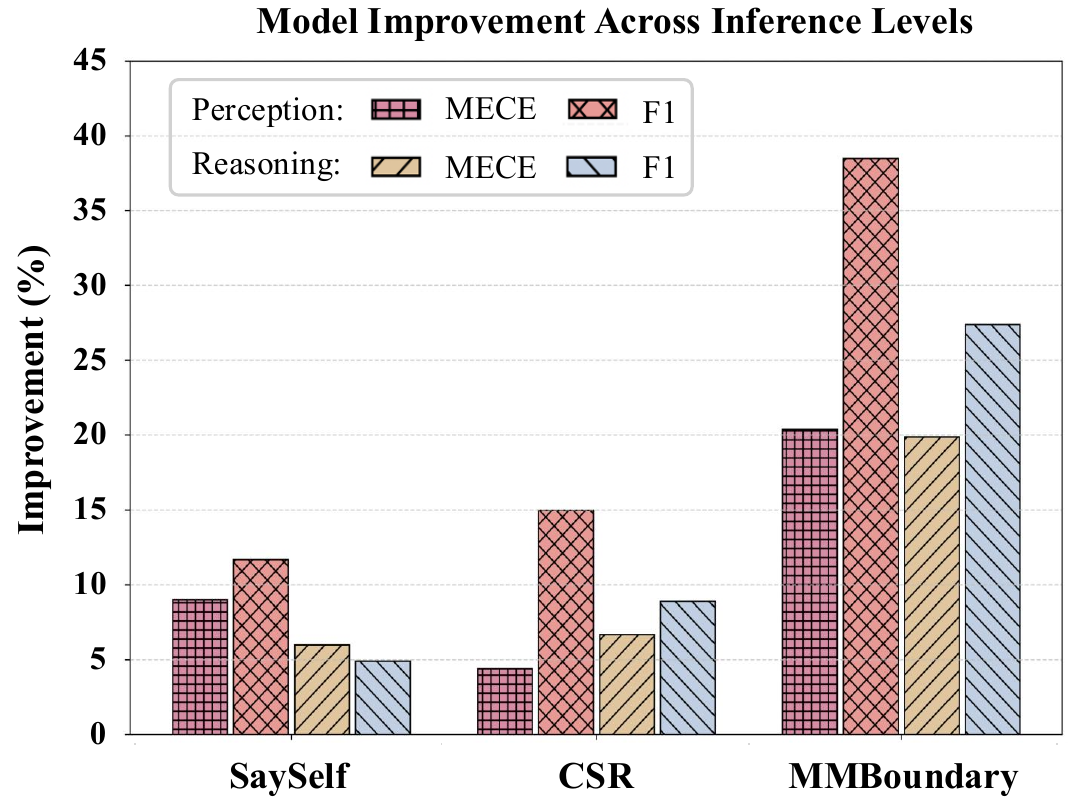}
  \caption{Performance improvement of strong baselines and our model compared to the base model in visual perception and cross-modal reasoning level of MLLMs. We report the results on ScienceVQA.}
  \label{fig:improve}
\end{figure}

\subsection{Effectiveness of Confidence Estimation}
\label{anaylsis_CE}
To evaluate the effectiveness of our proposed internal confidence estimation (ICE), we compare our method with the self-consistency-based confidence estimation method (SCE) \cite{yang2023alignment,xu2024sayself}. We randomly sample 50 data from three datasets and compare the confidence scores of the model’s responses from approaches above. 
We compute \( |C_{\text{ICE}} - C_{\text{SCE}}| \). The results are shown in Table \ref{tab:ICE}. We observe that the confidence estimation bias between the two methods is small for the vast majority of samples. On the ScienceVQA dataset, the average difference in confidence scores between the two methods is 0.0578, indicating that for a given answer from the model, our method has only about 6 instances of deviation compared to the post-hoc confidence estimation method. 
\subsection{Effectiveness of Confidence Calibration}
We further investigate the confidence calibration (MECE) and task performance (F1) of our method across different reasoning levels in MLLMs, specifically focusing on visual perception and cross-modal reasoning. The results is presented in Figure~\ref{fig:improve}. Our method achieves a significant improvement in confidence calibration at the perception level (an increase of 20.4\%), which contributes to a 38.5\% improvement in the accuracy of the reasoning chain. Furthermore, at the reasoning level, benefiting from the strengthened knowledge boundary in the visual understanding stage, both the confidence calibration score and the reasoning chain F1 score show improvements, surpassing the strongest baseline by 19.7\% and 27.4\%, respectively. 


%% file: sections/6_relatedwork.tex
\section{Related Work}
\noindent \textbf{Hallucinations and Uncertainty Estimation.} In MLLMs, hallucinations refer to model responses that are misaligned with the visual modality \cite{chen2024unified,liu2024survey,huang2025adactrladaptivecontrollablereasoning}, which can arise due to insufficient capabilities in visual perception and knowledge reasoning. 
Various efforts have been made to evaluate hallucination in MLLMs \cite{liu2024mitigatinghallucinationlargemultimodal, gunjal2024detecting, zhang2024knowledge,wu2024macaroon, xu2025amplifyadjacenttokendifferences}. As a fundamental approach to detecting model hallucination, uncertainty estimation (UE) has long attracted significant attention, falling into two main types: black-box and white-box. Black-box methods only require the generated text, and most of these methods are based on self-consistency \cite{fomicheva2020unsupervised,kuhn2023semantic,lin2023generating}, inspired by its success in textual domain reasoning \cite{wang2025calmunleashingcrosslingualselfaligning,he2024simucourt,jin2024rwku}. White-box methods rely on access to logits and internal layer outputs. They encompass information-based, density-based and sample diversity-based approaches \cite{malinin2020uncertainty, kadavath2022language, vazhentsev2023efficient, kuhn2023semantic, fadeeva2024fact, duan2024shifting}. 
Instead of simply relying on self-consistency prompting, we leverage the internal state of model to quantify the confidence of each reasoning step.

\noindent \textbf{Confidence Calibration of Language Model.} Existing research has highlighted the tendency of LLMs to fabricate information when faced with unknown questions \cite{hu2023won,amayuelas2023knowledge,liu2024examiningllmsuncertaintyexpression,huang2025mactuningllmmulticompositionalproblem,fan2025v2rbenchholisticallyevaluatinglvlm}. As a result, increasing attention has been directed toward enhancing the models' awareness of their knowledge boundaries and enabling them to express their confidence in outputs when encountering uncertainty \cite{lin2022teaching,xiong2023can, yang2023alignment, lyu2024calibrating, xu2024sayself, zhang2024r}. 
\citet{zhou2023navigating} empirically finds that injecting uncertainty expressions into prompts significantly increased the accuracy of GPT-3 responses and improved calibration scores. 
\citet{zhang2024r} introduce R-tuning to encourage LLMs to express "certain/not certain". \citet{xu2024sayself} goes further to teach the model to express more fine-grained confidence estimates along with self-reflective rationales. However, these methods focus solely on the entire response, which can lead to incorrect answers with high confidence. 
Therefore, we propose MMBoundary to train MLLMs to express fine-grained confidence estimates for each reasoning step, enhancing reasoning chain self-correction.

%% file: sections/7_conclusion.tex
\section{Conclusion}

In this work, we present MMBoundary, a novel framework that advances the knowledge boundary awareness of multimodal models through reasoning step confidence calibration. We incorporate complementary textual and cross-modal self-rewarding signals to estimate confidence at each step of the MLLM reasoning process. In addition to supervised fine-tuning MLLM for initial confidence expression warm-up, we further introduce a reinforcement learning stage with multiple reward functions for further calibrating model confidence. Empirical results demonstrate that our framework significantly outperforms existing methods, achieving an average reduction of 7.5\% in multimodal confidence calibration errors and up to 8.3\% improvement in task performance. 

%% file: sections/8_limitations.tex
\section*{Limitation}

Our framework aims to enable MLLMs to autonomously generate natural language confidence statements during inference, enhancing reasoning chain self-correction. A limitation of this work is that our proposed method involves using the model's internal states and uncertainty methods to assess the model's confidence. However, more research is needed to determine whether uncertainty methods can accurately reflect the model's confidence in its output. Ablation experiments on the uncertainty methods indicate that the four carefully selected methods provide gains for the model. Additionally, we explore the correlation between the proposed internal confidence estimation method and the self-consistency method. The results show that our metric, without requiring multiple samples, achieves performance comparable to methods that rely on multiple samples. 
To further advance our method, our future work will concentrate on the following areas: Firstly, we aim to incorporate multimodal self-correction mechanisms \cite{he2024selfcorrectionrefinementlearningframework}, leveraging diverse data sources to augment the model's capabilities. Secondly, we plan to explore synthetic data pre-training techniques \cite{qin2025scalinglawssyntheticdata} to address data scarcity and improve the model's generalization ability across various project scenarios. 

%% file: sections/appendix.tex
\appendix

\section{The Value-Statement Mapping Table}\label{sec:mapping_table}
This module aims to establish a mutual mapping between the detected score and predefined confidence statements. We set the confidence levels to five categories (uncertain, slightly uncertain, moderately confident, highly confident, and fully confident), considering that having more levels might lead to overly similar confidence statements between adjacent levels. The confidence statements need to be directly integrable into the model's generated content without sounding abrupt or redundant, much like human reflective expressions.
We have preset 40 concise statements for each level. Table \ref{tab:statements} presents additional confidence statements. These statements are concise and express the semantics of the corresponding confidence levels, allowing for seamless integration into sentences generated by the model, making them suitable for training generative language models.

\begin{table}[ht!]
\centering
\small
\begin{tabular}{c|p{5.5cm}}
\toprule
\multicolumn{1}{c}{\textbf{Score}} & \multicolumn{1}{c}{\textbf{Statement}}\\
\midrule
 1 & but I can't confirm this. / I'm uncertain about this. / I'm not sure about that. / This answer may be wrong. / I can't guarantee this answer. / I'm unsure about this. / I can't be sure about this. / This answer is unclear to me. / This might be imprecise. / This could be questionable. (…)
\\
\midrule
2  & and it may need checking / it might not be right. / but I'm not sure. / and it might be slightly off. / though it's not perfect. / but it may need confirmation. / though there's some doubt. / though it may not hold up. / though I feel a bit unsure. / but there's minor hesitation. (...)
\\
\midrule
3 & but I can't guarantee perfection. / I can't be entirely sure / but it's not beyond all doubt. / though minor errors might exist. / but it's not fully certain. / though small flaws are possible. / but it's not completely precise. / but it's not entirely error-free. / though it's not fully verified. / though it's open to review. (...)
 \\
\midrule
4  & and this seems trustworthy. / and I believe this is right. / and I'm quite confident in this. / and this feels reliable to me. / and I trust this is correct. / and this seems very likely true. / and this appears reliable. / and this fits the context well. / and I’m confident this is right. / and this is well-reasoned. (...)
 \\
\midrule
5  & with total certainty. / with no doubts at all. / and I'm absolutely sure about this. / and I'm fully confident in this. / with total certainty. / and this is undoubtedly correct. / and this is entirely reliable. / and it's unquestionably right. / with complete confidence. / and I guarantee this is right. (...)
\\
\bottomrule
\end{tabular}
\caption{\label{tab:statements}
The confidence score-statement mapping table. The five scores correspond to uncertain, slightly uncertain, moderately confident, highly confident, and fully confident. We preset 40 statements for each score.
}
\end{table}


\section{Implementation Details}
\label{sec:training}
Our experiment involves three distinct datasets: A-OKVQA \cite{schwenk2022okvqa}, ScienceVQA \cite{lu2022learn}, and CulturalVQA \cite{nayak2024benchmarking}. The first two datasets are in-domain datasets, and our training data comes from the training sets of these two datasets, while CulturalVQA is an out-of-domain dataset. Since the test sets for all three datasets are not publicly available, we cannot accurately annotate the reasoning chain for MECE and Reasoning Chain F1 evaluation. Therefore, we use the validation sets of AOKVQA and ScienceVQA for in-domain testing of the model. For CulturalVQA, which only has a non-public test set, we manually selected and annotated 800 samples from it to serve as the test set.

For the construction of the warm-up dataset, we deploy the vLLM model with a temperature setting of 0.1 and number of log probabilities to return per output token of 10. We collect a total of 19K Question-Image pairs from the training sets of A-OKVQA and ScienceVQA. For each Question-Image pair, we prompt the model to generate the reasoning chain and calculate the model's confidence score for each sentence, resulting in 55K sentences with confidence statements, with $w_0$, $w_1$, $w_2$ and $w_3$ all set to 0.25 in internal confidence estimation module. $\alpha$, $\beta$, $\gamma$ are equal. During the warm-up stage, we use the AdamW optimizer with a 10\% warm-up ratio, a learning rate of 1.0e-4, and a batch size of 16. In the reinforcement learning phase, we randomly sample data from the training set for training, for each question, we sample $N$ = 3, with a learning rate of 1e-5 and a batch size of 16. 

\begin{figure*}[!h]
  \includegraphics[width=1\textwidth]{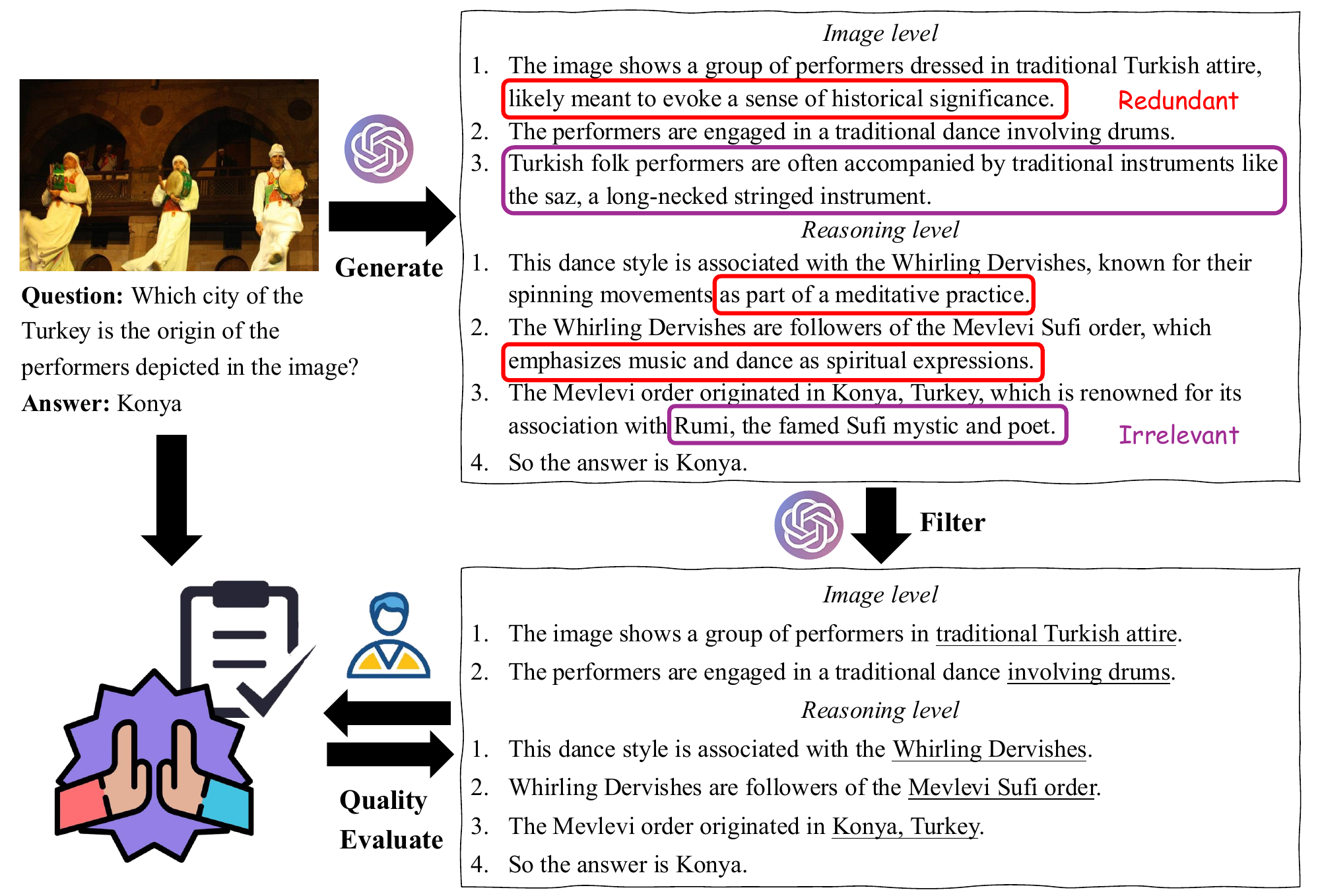}
  \caption{The Annotation Pipeline. We first prompt GPT-4o to generate an analysis (reasoning chain) structured at the perception and reasoning levels. Then, we have GPT-4o filter and correct the initially annotated chains. Finally, manual data quality control is conducted to ensure accuracy and reliability.}
  \label{fig:data_construct}
\end{figure*}

\section{The Reasoning Chain Annotation}
\label{sec:annotation}

To obtain the necessary fine-grained knowledge of visual perception and cross-modal reasoning in visual question-answering for calibrating the multi-level confidence of MLLMs, we conduct reasoning chain annotation on knowledge-extensive datasets from three different domains.

\subsection{The Annotation Pipeline}
The pipeline of reasoning chain annotation is presented in Figure~\ref{fig:data_construct}. We first prompt the GPT-4o to generate analysis (reasoning chain) structured in the perception and reasoning level. The former identifies key visual elements in the image that are most relevant to the question and answer, while the latter provides granularity reasoning that justifies why the answer is correct. Each level should include concise, interconnected sentences, with each sentence conveying a single piece of knowledge. As shown in the upper right corner of the figure, the initially obtained reasoning chain may contain redundant information and irrelevant content. Therefore, we use GPT-4o again to correct the content of the reasoning chain, filtering out redundancy and unrelated information to ensure that each sentence is concise and accurate. Then, we conduct annotation quality control to ensure the accuracy and consistency of the data. The prompt is provided in Appendix \ref{sec:prompt}. 

\subsection{Quality Evaluation}

After the machine annotation is completed, we randomly selected 50 samples from each of the three datasets and asked two graduate students to evaluate the data quality. The evaluation metrics included: (1)Accurate: the reasoning chain is relevant to the question and contains no wrong knowledge; (2) Concise: each sentence is concise and contains no redundant information; (3) Complete: the reasoning chain formed by each sentence accurately explains the answer to the corresponding question without omitting relevant knowledge. We use a Likert Scale to evaluate each indicator, with a scoring range from 1 to 5, where 1 indicates 'Strongly Disagree' and 5 indicates 'Strongly Agree.' The results are shown in Table \ref{tab:data_eval}. We report the proportion of data with a rating greater than 4 (i.e., Agree). The results indicate that the majority of the data meet the three criteria. We conduct a Kappa test on the accuracy evaluation results of the two graduate students, yielding a Kappa value of 0.75, which indicates a high level of consistency between the evaluators.

\begin{table}
\small
\centering
\begin{tabular}{c|ccc}
\toprule
Metric & Accurate & Concise & Complete  \\ \midrule
Rate (\%) & 96.8  & 91.3 &  93.5  \\
\bottomrule
\end{tabular}
\caption{The Likert Scale results of annotated data. We report the proportion of data with a rating greater than 4 (i.e., Agree).} 
\label{tab:data_eval}
\end{table}

\begin{figure*}[t]
  \centering
  \includegraphics[width=1\textwidth]{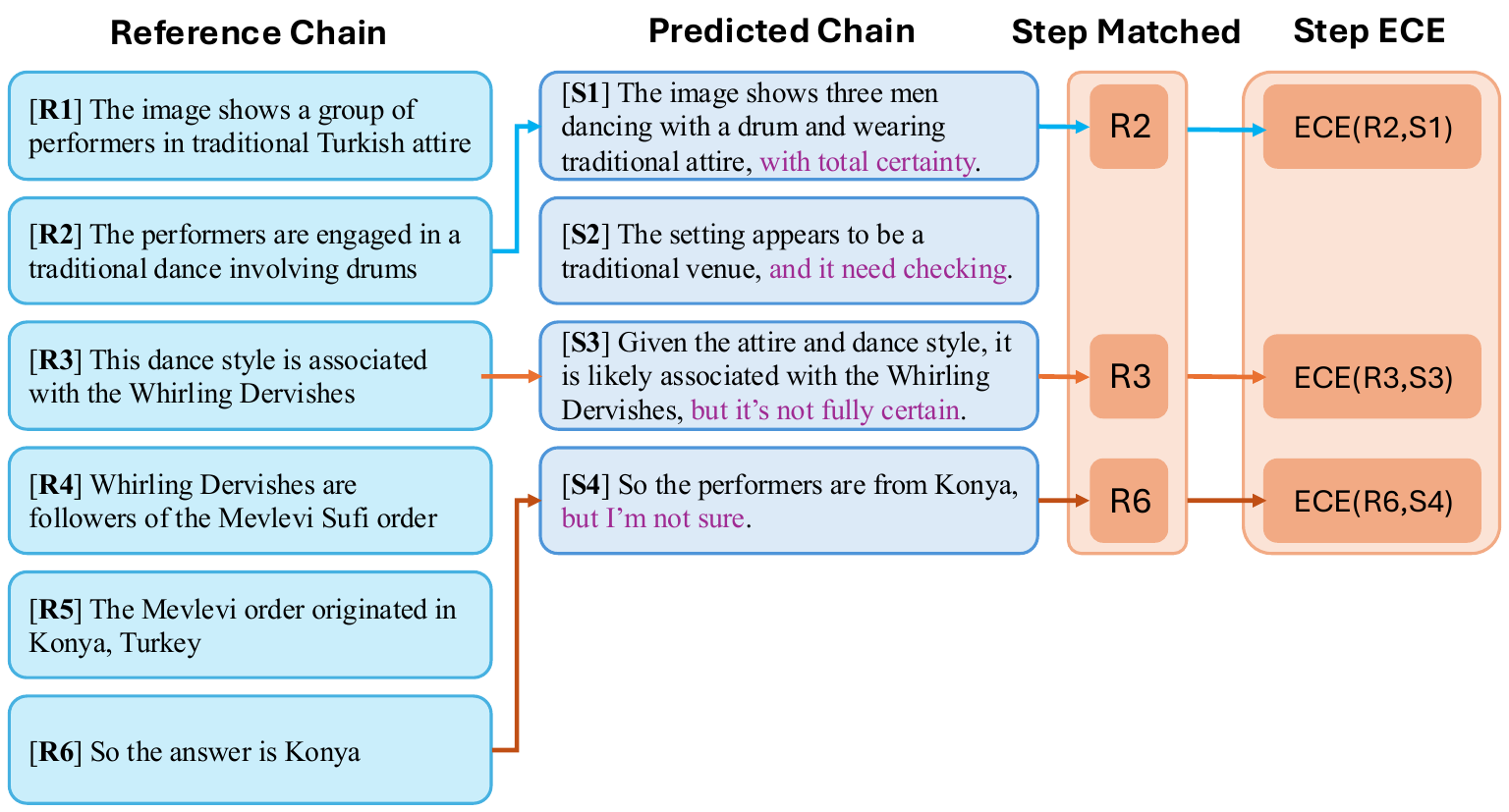}
  \caption{Example of MECE and Reasoning Chain F1 calculation.}
  \label{fig:MECE}
\end{figure*}

\begin{table*}[h!]
\centering
\resizebox{\linewidth}{!}{
\begin{tabular}{l|cccc|cccc|cccc}
\toprule
\multirow{2}{*}{\textbf{Model}} & \multicolumn{4}{c}{\textbf{A-OKVQA}} & \multicolumn{4}{c}{\textbf{ScienceVQA}} & \multicolumn{4}{c}{\textbf{CulturalVQA}}  \\ \cmidrule(lr){2-13} & \textbf{ECE} ($\downarrow$) & \textbf{MECE} ($\downarrow$) & \textbf{Acc} ($\uparrow$)  & \textbf{F1} ($\uparrow$) & \textbf{ECE} ($\downarrow$) & \textbf{MECE} ($\downarrow$) & \textbf{Acc} ($\uparrow$) & \textbf{F1} ($\uparrow$) & \textbf{ECE} ($\downarrow$) & \textbf{MECE} ($\downarrow$) & \textbf{Acc} ($\uparrow$) & \textbf{F1} ($\uparrow$)    \\ \midrule
Multisample & 0.437 & 0.463 & 0.665 & 0.557 & 0.467 & 0.492 & 0.613 & 0.415 & 0.443 & 0.512 & 0.335 & 0.513\\
SaySelf &0.324 & 0.381 & 0.727 & 0.632 & \textbf{0.332} & 0.445 & 0.628 & 0.523 & 0.397 & 0.426 & 0.352 & 0.586\\ 
Conf-CSR & 0.413 & 0.463 & 0.774 & 0.623 & 0.433 & 0.534 & 0.702 & 0.517 & 0.482 & 0.503 & 0.394 & 0.562 \\ 
RCE & 0.372 & 0.427 & 0.793 & 0.647 & 0.401 & 0.483 & 0.686 & 0.504 & 0.436 & 0.475 & 0.425 & 0.597\\ 
DRL &0.406 & 0.442 & 0.762 & 0.628 & 0.435 & 0.523 & 0.641 & 0.487 & 0.465 & 0.493 & 0.384 & 0.552 \\ 
\rowcolor{cyan!3.5}
MMBoundary & \textbf{0.305} & \textbf{0.348} & \textbf{0.806} & \textbf{0.664} & 0.346 & \textbf{0.426} & \textbf{0.713} & \textbf{0.562} & \textbf{0.354} & \textbf{0.385} & \textbf{0.437} & \textbf{0.634}\\ 
\bottomrule
\end{tabular}}
\caption{Experimental results on a different base model, Qwen2VL 7B \cite{wang2024qwen2}.}
\label{tab:results_other_base_model}
\end{table*}

\section{The Details of Evaluation Metrics}
\label{sec:metric}
\subsection{The Multi-granular Expected Calibration Error (MECE)}
\label{metric:MECE}
As shown in Figure \ref{fig:MECE}, after comparing the knowledge contained in the predictions and references, we obtain sentences where the knowledge in predictions and references aligns. Then, we calculate the Expected Calibration Error (ECE) for each sentence one by one, and finally derive the Multi-granular Expected Calibration Error (MECE):
\begin{equation}
\text{ECE}(a) = \frac{1}{|a|} \sum_{(z,c) \in a} \left| \mathbb{I}(z) - \text{Conf}(c) \right|
\end{equation}
\begin{equation}
\text{MECE}(A) = \frac{1}{|A|} \sum_{a \in A} \text{ECE}(a) 
\end{equation}
Here, $A$ represents the entire test set, and $a$ denotes the reasoning chain generated by the model, which consists of multiple sentences. $(z, c)$ represent a sentence and its corresponding confidence statement, respectively. $\mathbb{I}(\cdot)$ is the indicator function that returns 1 if the sentence is correct when compared with the reference chain, and 0 otherwise. $\text{Conf}(\cdot)$ represents the numerical value of the confidence statement.

\subsection{Area Under the Receiver Operating Characteristic curve (AUROC)}
\label{metric:auroc}
We adopt the \textit{\textbf{AUROC}} score \cite{hendrycks2016baseline}, which measures the ability of models to distinguish between correct and incorrect responses across different threshold settings. 
\begin{equation}
\text{AUROC} = \int_{0}^{1} \text{TPR}(\text{FPR}^{-1}(x)) \, dx \label{eq:auroc}
\end{equation}
where $x$ denotes the threshold confidence level, $TPR$ represents the true positive rate at this threshold, and $FPR$ indicates the false positive rate corresponding to the threshold. The result is shown in Table \ref{tab:results_auroc}.

\begin{table}[h!]
\small
\centering
\resizebox{\linewidth}{!}{
\begin{tabular}{l|cc|cc}
\toprule
\multirow{2}{*}{\textbf{Model}} & \multicolumn{2}{c}{\textbf{A-OKVQA}} & \multicolumn{2}{c}{\textbf{ScienceVQA}}   \\ \cmidrule(lr){2-5} & \textbf{ECE} ($\downarrow$) & \textbf{MECE} ($\downarrow$)  & \textbf{ECE} ($\downarrow$) & \textbf{MECE} ($\downarrow$)    \\ \midrule

\rowcolor{cyan!3.5}
MMBoundary & \textbf{0.316} & \textbf{0.324} & \textbf{0.354} & \textbf{0.405}\\
w/ UIC & 0.358 & 0.387 & 0.406 & 0.479 \\
\bottomrule
\end{tabular}}
\caption{The comparative study of confidence level segmentation methods. UIC (uniform interval for confidence level segmentation) means simply divides the interval [0, 1] directly into five equal segments, with each segment corresponding to a confidence level.}
\label{tab:uic}
\end{table}

\begin{figure}[t]
  \includegraphics[width=\columnwidth]{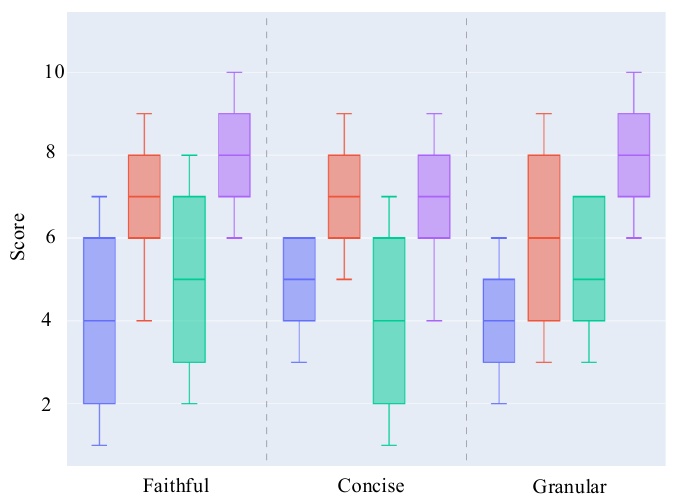}
  \caption{Boxplots of human evaluation scores on the A-OKVQA dataset.}
  \label{fig:eval_box}
\end{figure}

\subsection{Reasoning Chain F1 Score}
\label{metric:F1}
We use the Reasoning Chain F1 score \cite{ho2022wikiwhy} to evaluate the quality of the reasoning chain generated by the model. We compare the knowledge contained in the predictions and references. First, we split the predicted and reference chains into “steps” by sentence. We then compute a matrix of pairwise similarity scores before using a threshold to classify “matches.” Since a single predicted sentence may contain multiple reference knowledge, we keep separate counts of precise predicted sentences and covered reference sentences. These counts are then micro-averaged to calculate the overall precision, recall, and F1 scores for the test set:
\begin{equation}
\text{Precision} = \frac{\text{Matched}}{\text{Prediction}}, \text{Recall} = \frac{\text{Covered}}{\text{Reference}}  
\end{equation}
Taking the answer in Figure \ref{fig:MECE} as an example, we have: Prediction = 4, Reference = 6, Matched = 3, Covered = 3. We then calculate the F1 score:
\begin{equation}
F1 = 2 \times \frac{\text{Precision} \times \text{Recall}}{\text{Precision} + \text{Recall}}
\end{equation}
Drawing on the study of \citet{ho2022wikiwhy}, we select a large RoBERTa model (cross-encoder/stsb-roberta-large) with a similarity threshold of 0.64.

\section{Experiments of MMBoundary on Qwen}
\label{sec:extra_experiment}
To prove that our method can generalize on multiple models, we also implement the baseline approaches and MMBoundary on Qwen2VL 7B \cite{wang2024qwen2}.

\begin{table}[t]
\small
\centering
\resizebox{\linewidth}{!}{
\begin{tabular}{l|ccc}
\toprule
Model & A-OKVQA & Sci-VQA & Cul-VQA   \\ \midrule
Multisample & 0.5016& 0.5429& 0.4904 \\
SaySelf & \textbf{0.6872}& 0.6118& 0.6261 \\ 
Conf-CSR &0.5238&0.5713&0.4931 \\ 
RCE & 0.6037&0.5902&0.6059  \\ 
DRL & 0.4956&0.5028&0.4576 \\ 
\rowcolor{cyan!3.5}
MMBoundary & 0.6635& \textbf{0.6786}& \textbf{0.7108} \\ 
\bottomrule
\end{tabular}}
\caption{The AUROC experimental results.}
\label{tab:results_auroc}
\end{table}

\section{Analysis of Domain Shift}
We select three widely used datasets from distinct domains (general, scientific, and cultural) for separate training and testing. This cross-domain design aims to validate the robustness of our approach in domain-agnostic scenarios. The experimental results in Table \ref{tab:results_AE} and Table \ref{tab:results_other_base_model} demonstrate that: (1) Our confidence estimation method, which integrates textual and cross-modal signals based on the model's internal uncertainty, is inherently domain-agnostic, thereby effectively mitigating domain shift effects; (2) Despite notable distributional differences between domains, our method maintains a mean expected calibration error (MECE) of 0.361 on out-of-domain data, surpassing other strong baselines by over 7\%, demonstrating its good adaptability to domain shifts. 

\section{Prompt} 
\label{sec:prompt}
\subsection{Prompt for Data Annotation}
As illustrated in Figure \ref{fig:prompt1}, we present the prompt for data annotation. We first prompt the GPT-4o
to generate analysis (reasoning chain) structured
in the perception and reasoning level. The former
identifies key visual elements in the image that are
most relevant to the question and answer, while the
latter provides granularity reasoning that justifies
why the answer is correct.
\subsection{Prompt for Data Refinement}
The initially obtained reasoning chain may contain redundant information and irrelevant content. Therefore, we use GPT-4o again to correct the content of the reasoning chain, filtering out redundancy and unrelated information to ensure that each sentence is concise and accurate. As demonstrated in Figure \ref{fig:prompt2}, we present the refinement prompt designed to guide GPT-4o in filtering and correcting the initially annotated reasoning chains. 

\begin{figure*}[ht]
    \begin{tcolorbox}[colframe=blue!70!black, colback=blue!10!white, title=Prompt for Annotation, sharp corners]
        \begin{lstlisting}[style=plain]
You will receive a question, an accompanying image, the correct answer, and the corresponding rationales. Follow these steps to generate your analysis (reasoning chain) structured in two levels. 

Each level should include concise, interconnected sentences, with each sentence conveying a single piece of knowledge. Ensure the reasoning chain covers all necessary knowledge points concisely, with each sentence in this reasoning chain is essential and avoid adding redundant or irrelevant sentences.

The levels are as follows:
**Image level**: Identify key visual elements in the image that are mostly relevant to the question and answer. Format these sentences in JSON, like: ['sentence 1', ..., 'sentence i'].
**Reasoning level**: Based on the extracted visual elements, provide logical reasoning that justifies why the answer is correct. Format these in JSON as well: ['sentence i+1', 'sentence i+2', ..., 'So, the answer is ...'].

The sentences in both levels together should form a coherent chain of reasoning that clearly explains why the answer is correct. Ensure that each sentence builds upon the previous one to complete the reasoning chain. In the final sentence of the reasoning level, provide a clear conclusion with the answer, like: 'So, the answer is ...'.
 
Refer to the example below:
Please answer the following question:
Image: (Three people in traditional clothing holding drums, performing a form of the 'whirling dervishes' ritual.)
Question: Which city in Turkey is the origin of the performers depicted in the image?
Answer: Konya
Analysis: {{'Image_level': [], 'Reasoning_level': []}}
Your output:
Analysis: {{
    'Image_level': [
        'The image shows a group of performers in traditional Turkish attire',
        'The performers are engaged in a traditional dance involving drums'
    ],
    'Reasoning_level': [
        'This dance style is associated with the Whirling Dervishes',
        'Whirling Dervishes are followers of the Mevlevi Sufi order',
        'The Mevlevi order originated in Konya, Turkey',
        'So the answer is Konya'
    ]}}

Now, please answer the following question:
Image: image
Question: {question}
Answer: {answer}
Analysis:{{
    'Image_level': [],
    'Reasoning_level': []
    }}

Your output:
Analysis:
        \end{lstlisting}
    \end{tcolorbox}
    \caption{Prompt for data annotation. We first prompt the GPT-4o to generate analysis (reasoning chain) structured
in the perception and reasoning level.}
    \label{fig:prompt1}
\end{figure*}

\begin{figure*}[ht]
    \begin{tcolorbox}[colframe=blue!70!black, colback=blue!10!white, title=Prompt for data refinement, sharp corners]
        \begin{lstlisting}[style=plain]
Now, the following analysis (reasoning chain) is structured. 
Image_level and Reasoning_level together form a complete reasoning chain. Please filter out any irrelevant sentences to maintain a concise reasoning chain, including only the essential sentences.

Refer to the example below:
Image: (Three people in traditional clothing holding drums, performing a form of the 'whirling dervishes' ritual.)
Question: Which city in Turkey is the origin of the performers depicted in the image?
Answer: Konya
Analysis:{{
    'Image_level': [
        'The image shows a group of performers dressed in traditional Turkish attire, likely meant to evoke a sense of historical significance.',
        'The performers are engaged in a traditional dance involving drums.',
        'Turkish folk performers are often accompanied by traditional instruments like the saz, a long-necked stringed instrument.'
    ],
    'Reasoning_level': [
        'This dance style is associated with the Whirling Dervishes, known for their spinning movements as part of a meditative practice.',
        'The Whirling Dervishes are followers of the Mevlevi Sufi order, which emphasizes music and dance as spiritual expressions.',
        'The Mevlevi order originated in Konya, Turkey, which is renowned for its association with Rumi, the famed Sufi mystic and poet.',
        'So the answer is Konya.'
    ]}}

Your output:
Analysis:{{
    'Image_level': [
        'The image shows a group of performers in traditional Turkish attire.',
        'The performers are engaged in a traditional dance involving drums.'
    ],
    'Reasoning_level': [
        'This dance style is associated with the Whirling Dervishes.',
        'Whirling Dervishes are followers of the Mevlevi Sufi order.',
        'The Mevlevi order originated in Konya, Turkey.',
        'So the answer is Konya.'
    ]}}

Now:
Image: (image)
Question: {question}
Answer: {answer}
Analysis: {analysis}
Your output:
Analysis:
        \end{lstlisting}
    \end{tcolorbox}
    \caption{Prompt for data refinement. We use GPT-4o to correct the content of the reasoning chain, filtering out redundancy and unrelated information to ensure that each sentence is concise and accurate.}
    \label{fig:prompt2}
\end{figure*}